%% file: root.tex
\documentclass[letterpaper, 10 pt, conference]{ieeeconf}  

\IEEEoverridecommandlockouts                              

\usepackage{graphicx} 
\usepackage{amsmath} 

\usepackage{amsthm}
\usepackage{listings}
\theoremstyle{definition}

\usepackage{amssymb}
\usepackage{pifont}
\usepackage{booktabs}

\usepackage{algorithm}
\usepackage[noend]{algpseudocode}
\algnewcommand{\LineComment}[1]{\State \(\triangleright\) #1}
\usepackage[colorlinks=true]{hyperref}
\usepackage[font=scriptsize,labelfont=bf]{caption}
\usepackage{etoolbox}
\makeatletter
\patchcmd{\@makecaption}
  {\scshape}
  {}
  {}
  {}
\makeatother

\title{\LARGE \bf
Optimizing robot planning domains to reduce search time for long-horizon planning
}

\author{Maximilian Diehl$^{1}$, Chris Paxton$^{2}$, and Karinne Ramirez-Amaro$^{1}$ 
\thanks{$^{1}$Maximilian Diehl and Karinne Ramirez-Amaro. Faculty of Electrical Engineering, Chalmers University of Technology, SE-412 96 Gothenburg, Sweden.
        {\tt\small \{diehlm, karinne\}@chalmers.se}}%
\thanks{$^{2}$Chris Paxton is with NVIDIA, USA.
        {\tt\small cpaxton@nvidia.com}}%
}

\begin{document}

\maketitle
\thispagestyle{empty}
\pagestyle{empty}

\begin{abstract} 
We have recently introduced a system that automatically generates robotic planning operators from human demonstrations. One feature of our system is the operator count, which keeps track of the application frequency of every operator within the demonstrations. In this extended abstract, we show that we can use the count to slim down domains with the goal of decreasing the search time for long-horizon planning goals. The conceptual idea behind our approach is that we would like to prioritize operators that have occurred more often in the demonstrations over those that were not observed so frequently. We, therefore, propose to limit the domain only to the most popular operators. If this subset of operators is not sufficient to find a plan, we iteratively expand this subset of operators. We show that this significantly reduces the search time for long-horizon planning goals. 
\end{abstract}

\input{./Content/Complete.tex}

\addtolength{\textheight}{-12cm}   




\section*{Acknowledgment}
\addcontentsline{toc}{section}{Acknowledgment}
The research reported in this paper has been supported by Chalmers AI Research Centre (CHAIR).

\bibliography{mybib}
\bibliographystyle{IEEEtran}

\end{document}

%% file: Content/Complete.tex
\section{Introduction}
Search algorithms are the backbone of symbolic planning. Their task is to find a sequence of actions that allows a system to transit into a state that satisfies a set of goal conditions. However, the planning complexity can quickly grow with increasing goal distance. The scalability of plan generation to long-horizon tasks with a large number of objects is identified as an important problem in the planning community. Several countermeasures like hierarchical planning \cite{Kaelbling2011} or macro operator generation \cite{Botea2005, Chrpa2015, Hofmann2020} were proposed. Apart from the goal, however, also the domain complexity impacts the time that is required to generate plans. A planning domain provides blueprints of actions, also called planning operators, which define the available state transitions. Therefore, the more operator choices the planner has to consider for its search, the longer it will take to find a solution.

\begin{figure}[ht!]
\centering
  \includegraphics[width=0.48\textwidth]{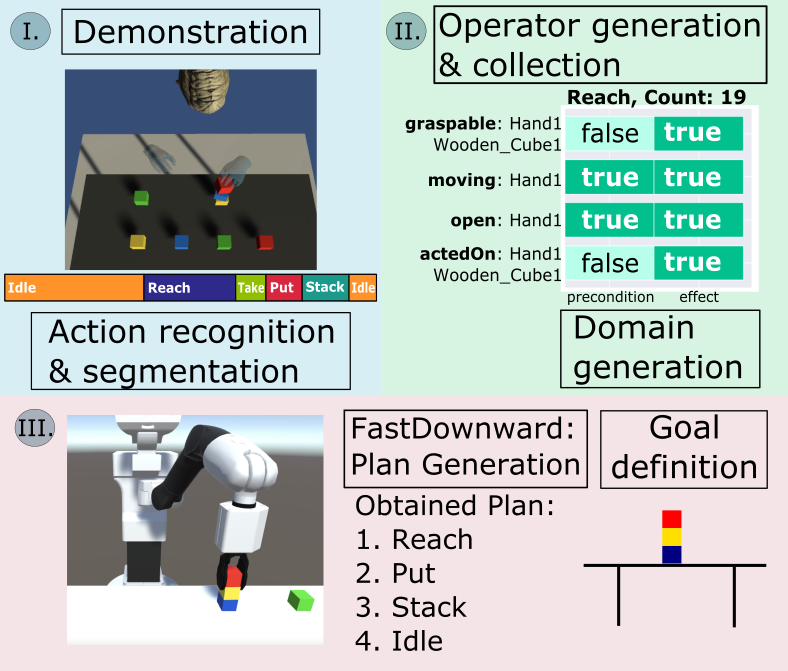}
  \caption{This figure illustrates the individual blocks of our system. I. Human demonstrations, performed in Virtual Reality, are segmented and classified. II. Planning operators are generated based on the classified activities and their preconditions and effects. At run-time, a planning domain is generated from the operator list. III. A plan is constructed with the help of off-the-shelf symbolic planners like Fast-Downward and executed by the TIAGo robot in a simulated environment.}
\label{fig:overview}
\vspace{-5mm}
\end{figure}

The operator and domain definition is usually done by hand, which is time-consuming or even requires a human expert~\cite{jimenez2012review}. We have, therefore, recently introduced a system that automatically generates robotic planning operators from demonstrations \cite{Diehl2021}. Furthermore, we have integrated the operator generation process with a plan generation and execution procedure, see Fig. \ref{fig:overview}. All collected operators are automatically parsed into a PDDL domain file. Plans are generated with the Fast-Downward planning system \cite{fastdownward} and executed on a simulated TIAGo robot. Our system keeps track of how often each of the operators occurred during the demonstrations through the operator count. 
This count helps us measure each operator's importance and can help the planning procedure guide the search. The next logical step is to investigate how the operator count could also be used to reduce the search complexity and thus speed up the plan generation.  

We therefore present a method that creates planning domains based on a subset of all collected planning operators. First, we will start analyzing only the most often observed operators and iteratively increase the number of operators in case that the plan search is not successful. We then analyze the impact of this approach in terms of its impact on the search time on the concrete example of the Fast-Downward planning system. The analysis is based on real operators generated through our system, as proposed in~\cite{Diehl2021}.

\section{Operator generation system and operator frequency measure}
Our system~\cite{Diehl2021} automatically generates planning operators from human demonstrations performed in a Virtual Reality setup. The first step of the system is the activity recognition, which is based on analyzing the hand movements of the demonstrators, as it was proposed in~\cite{BatesRIC17}. The demonstration is segmented and classified into high-level actions like \texttt{Reach, Take, Put, Stack or Idle}. The classification is based on a set of general rules in form of a C4.5 decision tree, which maps symbolic state variables like \texttt{inHand}, \texttt{actOn}, \texttt{handOpen}, \texttt{handMove} to the segmented activities. Since these parameters describe the state of the hand or gripper, we refer to them as \textit{hand-state variables}. Because of the generalisability of such semantic-based recognition methods, we could reuse the same set of rules (displayed in Table~\ref{tab:class}) from previous work~\cite{ramirez17AIJ}, which means that no training was required in our application.
\begin{table}[h]
\begin{center}
\caption{\label{tab:class}Hand activity classification rules. T and F stand for true and false respectively, and $\neg$nil means an object as opposed to no-object (nil).}
\begin{tabular}{| c | c | c | c | c | c |}
\hline
 features & \textit{Stack} & \textit{Idle} & \textit{Reach} & \textit{Put} & \textit{Take} \\ 
  \hline
  \hline
 {\tt handMove} & T & T $\lor$ F & T & T & F\\
 \hline
 {\tt actedOn} & $\neg$nil & nil & $\neg$nil &  nil & nil\\
 \hline
 {\tt inHand} & $\neg$nil & nil & nil & $\neg$nil & $\neg$nil\\
  \hline
\end{tabular}
\vspace{-3mm}
\end{center}
\end{table}

The system utilizes the segmentation to create new operators and its classification for a meaningful operator name, but additionally, also all the relevant preconditions and effects need to be extracted. For the state description, we re-used the set of \textit{hand-state variables} which we already required for the activity segmentation and classification (\texttt{inHand}, \texttt{actOn}, \texttt{handOpen}, \texttt{handMove}). On top of that, we added three commonly used state variables, \texttt{graspable}, \texttt{onTop} and \texttt{inTouch}. \texttt{Graspable} describes the situation where the hand is in close range to an object. Besides the \textit{hand-state variables}, the operators should reflect changes in the environment as well. In particular, the hand activities need to be mapped to all the respectively caused environment changes (e.g., in terms of \textit{environment-state variables} like \texttt{onTop} and \texttt{inTouch}). The resulting set of state variables (displayed with respective groundings in Table~\ref{tab:stateVar}) is minimalistic, but nevertheless expressive enough for representing the cube stacking environment on a high, symbolic level.  

\begin{table}[h!]
\begin{center}
\caption{Shows the defined state variables (\texttt{sv}), and their respective grounding during demonstration and execution. The object categories inside the brackets after the sv-name indicate the range. The thresholds were determined heuristically.}
\label{tab:stateVar}
\begin{tabular}{ l l }
\toprule
 State variables (${\tt sv}$) & Grounding \\ 
 \midrule
  \multicolumn{2}{ l }{\textbf{Hand --} ${\tt sv}$} \\
${\tt inHand}(\texttt{Hand},\texttt{Cube})$ & A hand/gripper has closed its fingers \\
& around a cube. \\
${\tt actedOn}(\texttt{Hand},\texttt{Cube})$ &
Dist. betw. cube and hand $<0.16$m. \\
& \& hand moving towards cube. \\
${\tt handMove}(\texttt{Hand})$ & 
Hand is moving $>0.1 \text{m}/\text{s}$ \\
${\tt graspable}(\texttt{Hand},\texttt{Cube})$ & 
Dist. betw. cube and hand $<0.1$m.\\
${\tt handOpen}(\texttt{Hand})$ &
Hand is open.\\
\midrule
\multicolumn{2}{ l }{\textbf{Environment --} ${\tt sv}$} \\
${\tt inTouch}(\texttt{Thing},\texttt{Thing})$ & 
VR physics engine detects contact \\
& between 2 objects. \\
${\tt onTop}(\texttt{Thing},\texttt{Thing})$ & 
Obj. A on top of obj. B if inTouch and \\
& A higher than B. \\  
\bottomrule
\end{tabular}
\end{center}
\vspace{-3mm}
\end{table}

Our system iteratively collects operators with each new demonstration. If the operator is new in terms of its preconditions and effects, it is added to the list of operators. If the operator has already been encountered on a prior occasion, the operator counter is incremented. The system uses the operator frequency as a quality measure. The goal is to prefer more often observed operators for the following two reasons. First of all, are more often observed operators less susceptible to potential segmentation errors. Second, they are the preferred choice of the human and should therefore be considered as an essential measure for the plan generation. Since planning systems like Fast-Downward \cite{fastdownward} perform minimization, we introduced a score inverse proportional to the operator count: 
\begin{equation*}
    \text{Cost}(op) = \lceil \lambda(1 - \frac{op_{count}}{op\_type_{count}}) \rceil,
\end{equation*}
with $\lambda = 100$.

\begin{figure}[ht!]
\centering
  \includegraphics[width=0.48\textwidth]{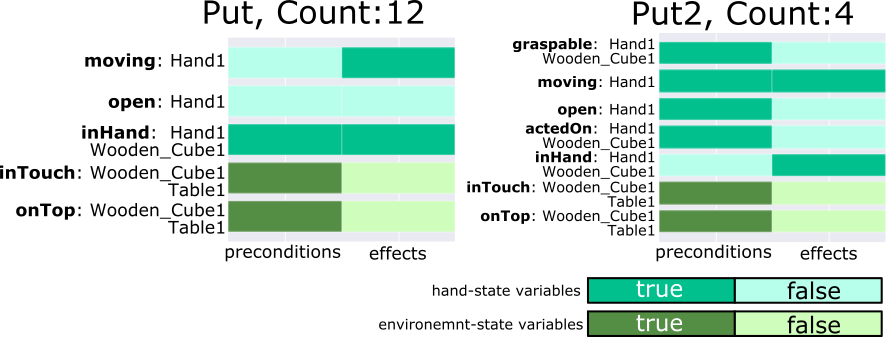}
  \caption{The two most often observed configurations of the \texttt{Put} operator.}
\label{fig:put}
\end{figure}

In \cite{Diehl2021}, we have tested the operator generation system based on 12 demonstrations from 3 different participants, on the task of stacking one or two cubes with the left or right hand. Our system currently classifies operators into the five different categories/types of \texttt{Reach, Take, Put, Stack} and \texttt{Idle}. For each of the operator types, several different configurations in terms of preconditions and effects were observed. In total, 115 activities were retrieved from the data, producing 30 unique operator configurations distributed among the five categories. The system, for example, generated five different \texttt{Put} operators, of which the two most often observed configurations are displayed in Figure~\ref{fig:put}. 
In most cases, the user first \textit{reaches} for a cube, \textit{takes} and then \textit{puts} it somewhere. This yields the (Put, Count: 12) operator.  However, occasionally (Put2, Count: 4), the user was picking up objects without resting the hand, which is one of the decision rules for \textit{Take} (see Table~\ref{tab:class}). Thus there was a direct transition from \textit{Reach} to \textit{Put}. Note, for example, that the \texttt{inHand} predicate (5th row) is changing from false to true for \texttt{Put2}, whereas it must already be satisfied as a precondition (3rd row) for \texttt{Put}. 

\section{Domain optimization}
With increasingly longer planning horizons, finding a plan takes substantially more time. For example, finding a plan to build a tower of six cubes takes more than 180 seconds with A* search and iPDB heuristic and even more than 1725 seconds with A* search and landmark-cut heuristic.

One way of improving plan generation performance would be to reduce the number of planning operators. We, therefore, propose a method to slim down our planning domains to operators who were observed most often and iteratively increase the scope to less popular operators in case no plan can be found on the more exclusive subset of operators. 

The exact procedure of the domain optimization works as described in Algorithm \ref{alg:slimOps}. The variable $\textit{prio}$ denotes how many operators of each type should be included in the domain: for $\textit{prio} = 1$, only the operators with the highest count, for $\textit{prio} = 2$, the first and second most frequent operators and so on. If there is a tie among the demonstration frequency of several operators of the same type, all of these operators are included. 

\setlength{\textfloatsep}{2pt}
\begin{algorithm}
  \caption{\label{alg:slimOps}Iterative expansion of domain operators}
  \hspace*{\algorithmicindent} \textbf{Input:} list of operators $O$, goal specification $g$ \\
  \hspace*{\algorithmicindent} \textbf{Output:} plan $\pi$ to reach the goal
  \begin{algorithmic}[1]
  \State $\textit{plan\_gen\_succ} \leftarrow 0$ \Comment{plan generation success} 
  \State $\textit{prio} \leftarrow 1$ \Comment{operator priority}
  \While{$\textit{plan\_gen\_succ} == 0$} 
   \State $\textit{O\_subs} \leftarrow []$
   \ForAll{$op\_type$} 
   \LineComment{Add most used operators for each operator type.} 
   \State $\textit{O\_subs}.append(\textsc{OperatorSubset}(O, \textit{prio}))$
   \EndFor
   \State $\textit{domain} \leftarrow \textsc{GeneratePDDLDomain}(\textit{O\_subs})$
   \State $\pi, \textit{plan\_gen\_succ} \leftarrow \textsc{GeneratePlan}(\textit{domain}, g)$
   \State $\textit{prio} \leftarrow \textit{prio} + 1$
  \EndWhile
  \end{algorithmic}
\end{algorithm}

\section{Experiment}
\begin{figure}[ht!]
\centering
  \includegraphics[width=0.4\textwidth]{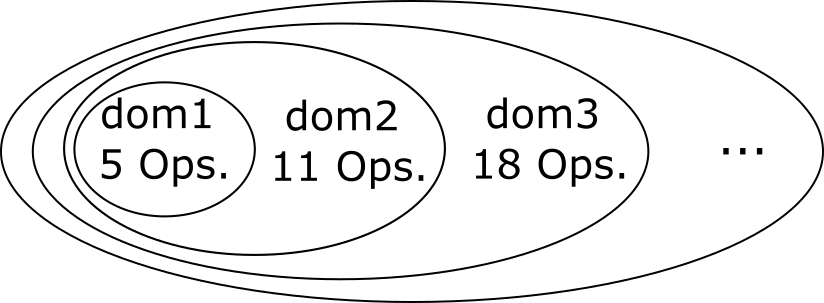}
  \caption{Iteratively increasing set of operators. }
\label{fig:op_bag}
\vspace{-3mm}
\end{figure}

\begin{figure}[ht!]
    \centering
      \includegraphics[width=0.48\textwidth]{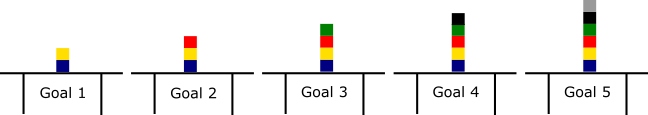}
      \caption{Test set of 5 increasingly difficult planning goals.}
    \label{fig:planning_goals}
\end{figure}

To test the impact of considering different sets of operators we have conducted several experiments. First, we use the same set of 12 demonstrations from our user experiment in \cite{Diehl2021}. Based on our proposed method, we generate in total five different domains with an increasing number (5, 11, 18, 23, 30 respectively; see Figure \ref{fig:op_bag}) of operators from this data. We then evaluate the search time for each domain on a set of increasingly difficult planning problems (Figure \ref{fig:planning_goals}).


\begin{figure}[ht!]
    \centering
      \includegraphics[width=0.45\textwidth]{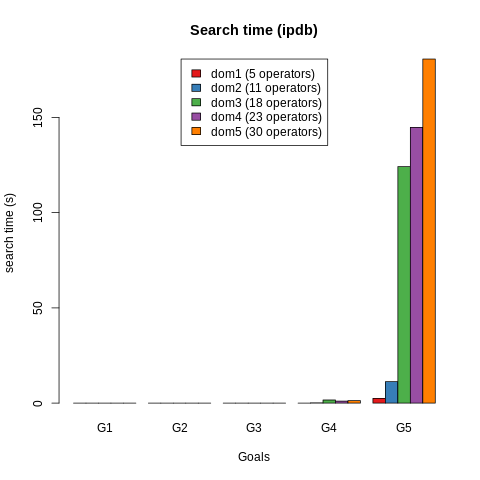}
    \caption{Planning time of the A* search with iPDB heuristic, plotted for 5 different stacking goals with increasing complexity for the 5 created domains.} 
    \label{fig:search_time_ipdb}
    \vspace{-5mm}
\end{figure}

\begin{figure}[ht!]
    \centering
      \includegraphics[width=0.45\textwidth]{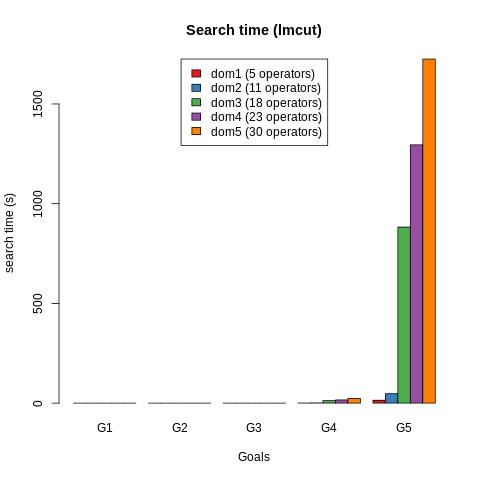}
      \caption{Planning time of the A* search with landmark-cut heuristic, plotted for 5 different stacking goals with increasing complexity for the 5 created domains.}
    \label{fig:search_time_lmcut}
    \vspace{-5mm}
\end{figure}

Our domains require planning systems to support the \textit{action-cost:} feature from PDDL3.1. We, therefore, chose the Fast-Downward planning system~\cite{fastdownward}, which offers a wide variety of different heuristics and planning algorithms. For the purpose of our experiment we use the two configurations of \textit{astar(ipdb())} (A* search with iPDB heuristic) and \textit{astar(lmcut())} (A* search with landmark-cut heuristic). Note that the purpose of testing two different heuristics is to show that the search time reduction, achieved by slimming the domains, is independent of the planning system. 
Results are displayed in Figures~\ref{fig:search_time_ipdb} and \ref{fig:search_time_lmcut}.

\begin{table}[ht!]
\begin{center}
\caption{\label{tab:times_ipdb}Search time (in seconds) for each of the 5 domains for the 5 planning goals as visualised in Figure \ref{fig:planning_goals} for the iPDB heuristic.}
\begin{tabular}{| c | c | c | c | c | c |}
\hline
 domain & Goal 1 & Goal 2 & Goal 3 & Goal 4 & Goal 5 \\ 
  \hline
  \hline
 {\tt dom1(5)} & 0.0005 &	0.00066  &	0.0011 &	0.0013 &	2.499
\\
 \hline
 {\tt dom2(11)} & 0.00042&	0.00065 &	0.0011 &	0.1626 &	11.2901
\\
 \hline
 {\tt dom3(18)} & 0.0005 &	0.00061 &	0.0011 &	1.6405 &	124.2
\\
 \hline
 {\tt dom4(23)} & 0.00042 &	0.00064 &	0.0011 &	1.0601 &	144.769
\\
 \hline
 {\tt dom5(30)} & 0.0005 &	0.00093 &	0.00081 &	1.2987 &	180.652
\\
  \hline
\end{tabular}
\end{center}
\end{table}
\begin{table}[ht!]
\begin{center}
\caption{\label{tab:times_lmcut}Search time (in seconds) for each of the 5 domains for the 5 planning goals as visualised in Figure \ref{fig:planning_goals} for the landmark-cut heuristic.}
\begin{tabular}{| c | c | c | c | c | c |}
\hline
 domain & Goal 1 & Goal 2 & Goal 3 & Goal 4 & Goal 5 \\ 
  \hline
  \hline
 {\tt dom1(5)} & 0.00048 &	0.0015& 	0.0046& 	0.2877& 	14.30

\\
 \hline
 {\tt dom2(11)} & 0.00085& 	0.0028& 	0.0103& 	0.977& 	47.415

\\
 \hline
 {\tt dom3(18)} & 0.00072& 	0.0093& 	0.0751& 	13.081& 	882.75

\\
 \hline
 {\tt dom4(23)} & 0.0008& 	0.01135& 	0.0890& 	15.96& 	1295.11

\\
 \hline
 {\tt dom5(30)} & 0.0018& 	0.01348	& 0.1588& 	23.186	& 1725.18

\\
  \hline
\end{tabular}
\end{center}
\end{table}

First off, we can see that for all test goals of this particular experiment, the specific set of 5 operators (the one most often demonstrated operator for each operator type) is enough to produce a plan. We also observe from tables \ref{tab:times_ipdb} and \ref{tab:times_lmcut}, that the complexity of the first three goals is not large enough to detect significant differences between the domains in terms of search time. In particular, the minimal operator domain is not necessarily the fastest. However, the difference is significant for the fifth planning goal, which takes at least 20 consecutive actions to reach. The search takes around 72 times (2.499s vs. 180.652s) more time for the iPDB heuristic and even 120 times (14.3s vs. 1725.18s) for the landmark-cut heuristic comparing the smallest and the largest domain.
\section{Discussion}
The experiment clearly shows significant gains in search time reduction for complex goals when slimming the domain to only the most often used operators. Note that, in general, there is no guarantee that the first batch of operators always finds a plan. Nevertheless, for long-horizon plans, the difference in search time is so significant that we would save time even if the first domains fail to generate any plans, and the scope of operators would need to be extended one or two times. Additionally, this gap in search time will grow even larger with more complex plans.

Another advantage is that reducing the number of operators can also force the planning system to follow the consensus of the human demonstration more closely, rather than using loopholes for optimization purposes. Consider our \texttt{Put} operators (see Figure~\ref{fig:put}) as an example. A plan like [\texttt{Reach}, \texttt{Put2}, \texttt{Stack}] is cheaper than [\texttt{Reach}, \texttt{Take}, \texttt{Put}, \texttt{Stack}], because the cost of \texttt{Take} + \texttt{Put} is higher than just \texttt{Put2}, even if \texttt{Put2} was less often observed than \texttt{Put}. In the first domain with just 5 operators, however, only \texttt{Put} is available, which results in a plan that follows more closely what most users were demonstrating. 

While the domain slimming method has many upsides in the discussed example, there are also some potential challenges. When scaling up the domain by including operators from different demonstration scenarios, there could be a bias towards environments where the number of observations was large. For example, compare the cube stacking experiment with 12 demonstrations with a potential dish stacking experiment with 100 demonstrations. The resulting operators will have a much higher count and therefore be added earlier into the subset of considered operators. That will happen even for applications where there are no dishes at all. Consequently, it will also be essential to work on additional operator merging techniques, like operator generalization over several object types. Generalization would possibly allow extending application environments and tasks. However, its success might depend on how detailed state variables describe the world state. If, for example, we introduce additional state variables which closely describe the shape of objects, the resulting operator will also be more specific and less generalizable. Essentially it is a question of what an appropriate level of abstraction looks like and could also require a hierarchical approach. 

\section{Conclusion}
In this paper, we have investigated the problem of reducing search time for long-horizon planning goals. We present a method that creates planning domains based on a subset of all collected planning operators. First, we will start analyzing only the most often observed operators and iteratively increase the number of operators if the plan search is unsuccessful. We showed that this significantly reduces the planning time.
As a next step, we would like to work more on merging operators, for example, by generalizing over object types. This could go hand in hand with the transition to probabilistic operators and will also comprise the interesting investigation of 
how to balance operator expressiveness vs. generalizability in terms of abstraction and choice of state variables.